\documentclass[11pt,a4paper,bookmarks=false]{article}
\usepackage{emnlp2020}
\usepackage{times}
\usepackage{latexsym}
\usepackage{graphicx}
\usepackage{amsmath}
\usepackage{amsfonts}
\usepackage{multirow}
\usepackage[linesnumbered, boxed, ruled]{algorithm2e}

\SetKwProg{Fn}{Function}{\string:}{end}

\usepackage{microtype}
\newcommand{\secref}[1]{Section \ref{#1}}
\newcommand{\figref}[1]{Figure \ref{#1}}

\newcommand{\cut}[1]{}

\aclfinalcopy 

\title{ST$^2$: Small-data Text Style Transfer via Multi-task Meta-Learning}

\author{Xiwen Chen, Kenny Q. Zhu \\
  Advanced Data and Programming Technologies Lab \\
  Shanghai Jiao Tong University \\
  \texttt{\{victoria-x@sjtu, kzhu@cs.sjtu\}.edu.cn} \\
}

\date{}

\begin{document}
\maketitle
\begin{abstract}

Text style transfer aims to paraphrase a sentence in one style 
into another style while preserving content. Due to lack of parallel training
data, state-of-art methods are unsupervised and rely on large datasets
that share content. Furthermore, existing methods have been applied on 
very limited categories of styles such as positive/negative and formal/informal. 
In this work, we develop a meta-learning framework 
to transfer between any kind of 
text styles, including personal writing styles that are more 
fine-grained, share less content and have much smaller training data.
While state-of-art models fail in the few-shot style transfer task, 
our framework effectively utilizes information from other styles to 
improve both language fluency and style transfer accuracy. 

\end{abstract}

\section{Introduction}
\label{sec:intro}

Text style transfer aims at rephrasing a given sentence in a desired style. It can be used to rewrite stylized literature works, generate different styles of journals or news (e.g., formal/informal), and to transfer educational texts with specialized knowledge for education with different levels.

Due to lack of parallel data for this task, 
previous works mainly focused on unsupervised learning of styles, 
usually assuming that there is a substantial amount of nonparallel 
corpora for each style, and that the contents of the two corpora do not 
differ significantly~\cite{shen2017style,john2018disentangled,fu2018style}. 
Existing state-of-art models either attempt to disentangle style and 
content in the latent 
space~\cite{shen2017style,john2018disentangled,fu2018style}, 
directly modifies the input sentence to remove stylized words~\cite{li2018delete}, 
or use reinforcement learning to control the generation of 
transferred sentences in terms of style and 
content~\cite{wu2019hierarchical,luo2019dual}. 
However, most of the approaches fail on low-resource datasets based on 
our experiments. This calls for new few-shot style transfer techniques. 


The general notion of style is not restricted to the heavily studied 
sentiment styles, but also writing styles of a person. However, even the
most productive writer can't produce a fraction of the text corpora 
commonly used for unsupervised training of style transfer today.
Meanwhile, in real world, there exists as many writing styles as you can
imagine. Viewing the transfer between each pair of styles as a 
separate domain-specific task, we can thus formulate
a multi-task learning problem, each task corresponding to a pair of styles. 
To this end, we apply a meta-learning scheme to take advantage of
data from other domains, i.e., other styles to enhance the performance 
of few-shot style transfer~\cite{finn2017model}.

Moreover, existing works mainly focus on a very limited range of styles. 
In this work, we take both personal writing styles and previously studied 
general styles, such as sentiment style, into account. 
We test our model and other state-of-the-art style transfer models on two datasets, 
each with several style transfer tasks with small training data, and verify 
that information from different style domains used by our model enhances 
the abilities in content preservation, style transfer accuracy, 
and language fluency. 

Our contributions are listed as follows:
\begin{itemize}
	\item We show that existing state-of-the-art style transfer models fail
	on small training data which naturally shares less content (see \secref{sec:st}
	and \secref{sec:pretrain}).
	\item We propose Multi-task Small-data Text Style Transfer (ST$^2$) algorithm, 
	which adapts meta-learning framework to existing state-of-art models, 
	and this is the first work that applies meta-learning on text style transfer 
	to the best of our knowledge (see \secref{sec:approach}).
	\item The proposed algorithm substantially outperforms the 
	state-of-the-art models in the few-shot text style transfer in terms of 
	content preservation, transfer accuracy and language fluency 
	(see \secref{sec:eval}).
	\item We create and release a literature writing style transfer
	dataset, which the first of its kind (see \secref{sec:lt}).
\end{itemize}

\section{Approach}
\label{sec:approach}
In this section, we first present two simple but effective style transfer models, namely \emph{CrossAlign}~\cite{shen2017style} and \emph{VAE}~\cite{john2018disentangled} as our base models, and then present a meta-learning framework called model-agnostic meta-learning~\cite{finn2017model} that incorporates the base models to solve the few-shot style transfer problem.

\subsection{Preliminaries}
%
\subsubsection*{Cross Align}

\begin{figure}[htbp]
	\centering
	\includegraphics[width=7cm]{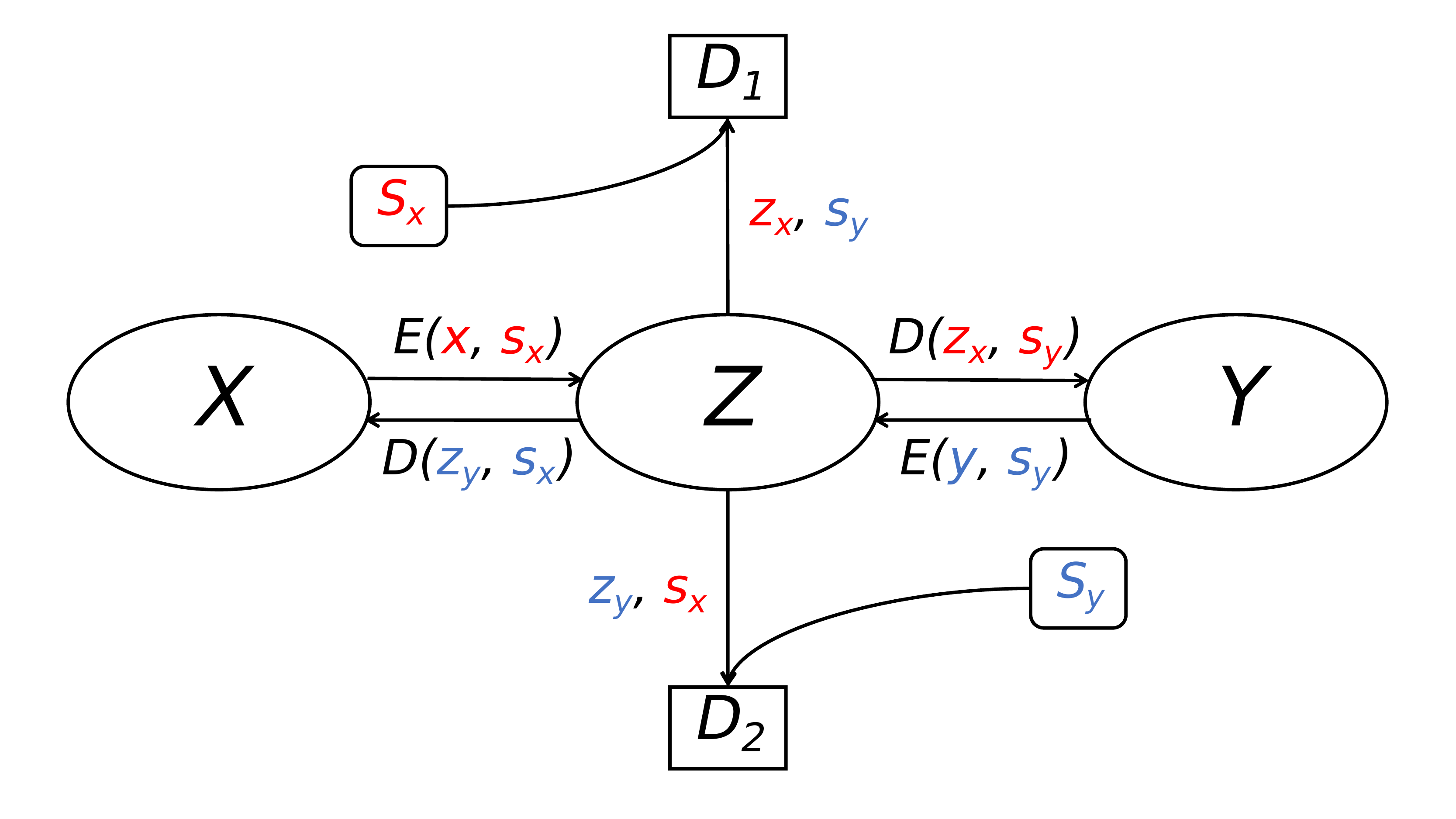}
	\caption{CrossAlign architecture}
	\label{fig:crossalign}
\end{figure}

The \emph{CrossAlign} model architecture proposed by \citet{shen2017style} is shown in \figref{fig:crossalign}. Let $X$ and $Y$ be two corpora with styles $s_x$ and $s_y$, respectively. $E$ and $D$ are encoders and decoders that take both the sentence $x$ or $y$, and their corresponding style labels $s_x$ or $s_y$ as inputs. Then the encoded sentences $z_x$ and $z_y$, together with their labels are input to two different adversarial discriminators $D_1$ and $D_2$, which are trained to differentiate between logits generated by the concatenation of content embedding and the original/opposite style label.

In training phase, the discriminators and the seq2seq model are trained jointly. The objective is to find
\begin{align*}
\theta^* = \underset{\theta}{\arg\min\ } \mathcal{L}_{\mathrm{rec}} (\theta_E, \theta_D) + \mathcal{L} _{\mathrm{adv}}(\theta_E, \theta_D),
\end{align*}
where
\begin{align*}
\mathcal{L}_{\mathrm{adv}}(\theta_E, \theta_D) & = \mathbb{E}_{x\sim X}[-\log D(E(x, s_x))] & \\
& + \mathbb{E}_{y\sim Y}[-\log (1 - D(E(y, s_y)))].
\end{align*}
The discriminators are implemented as CNN classifiers~\cite{kim2014convolutional}.

\subsubsection*{VAE for Style Transfer}

In order to disentangle style and content in the latent space, \citet{john2018disentangled} used variational autoencoder (VAE) and their specially designed style-oriented and content-oriented losses to guide the updates of the latent space distributions for the two components~\cite{kingma2013auto}. 

\begin{figure}[htbp]
	\centering
	\includegraphics[width=7cm]{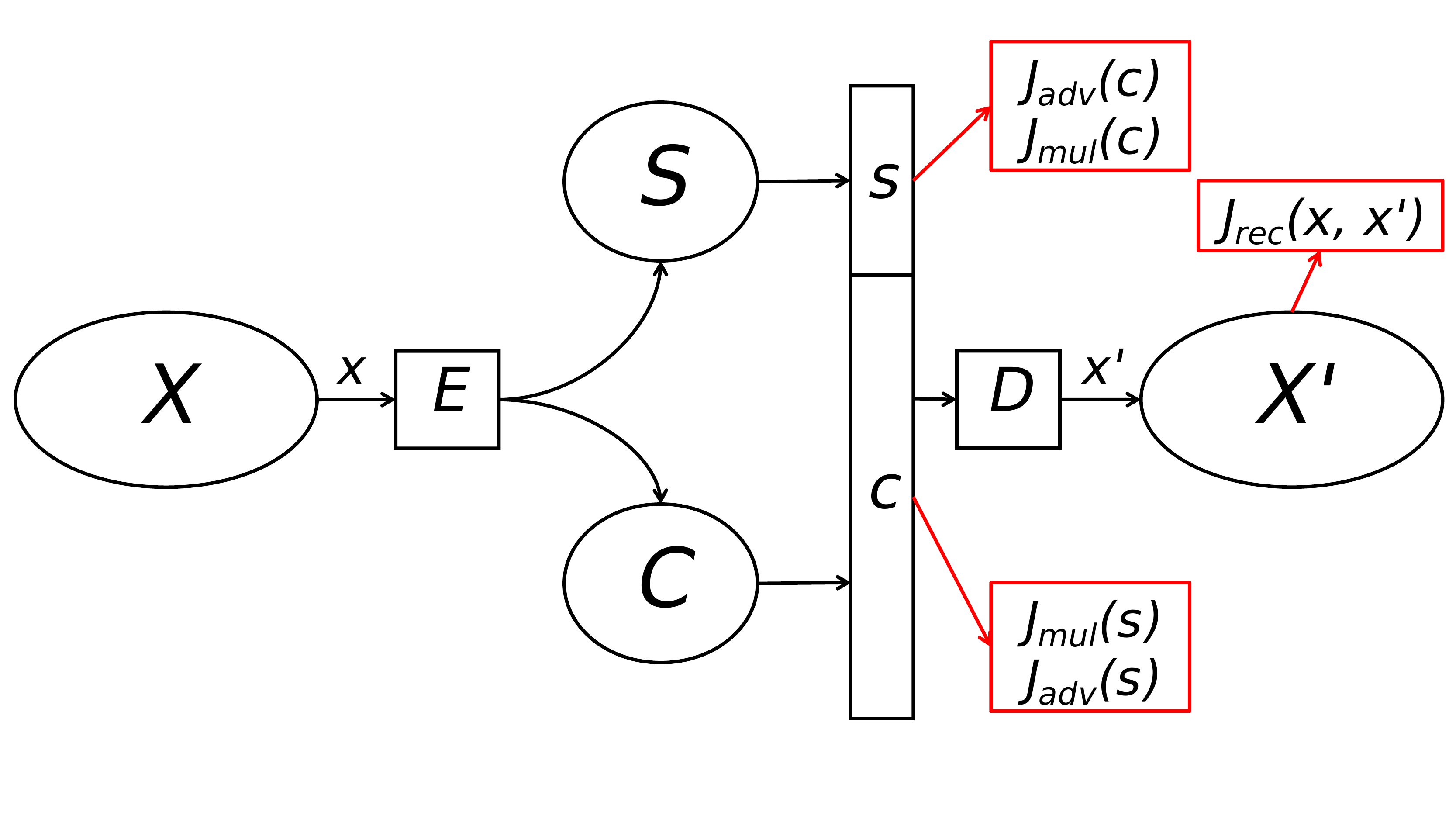}
	\caption{VAE architecture}
	\label{fig:vae}
\end{figure}

The architecture of this model is shown in \figref{fig:vae}. Given a corpus $X$ with unknown latent style space and content space, 
an RNN encoder maps a sequence $x$ into the latent space, 
which defines a distribution of style and content~\cite{cho2014learning}. 
Then style embedding and content embedding are sampled from 
their corresponding latent distributions and are concatenated 
as the training sentence embedding. 

The two embeddings are used to calculate multi-task 
loss $J_{\mathrm{mul}}$ and adversarial loss $J_{\mathrm{adv}}$ 
for content and style to separate their information. 
Then this concatenated latent vector is used as a generative 
latent vector, and is concatenated to every step of the input sequence
and fed into decoder $D$, which reconstructs the sentence $x'$. 
The final loss is the sum of these multi-task losses and the usual VAE reconstruction $J_{\mathrm{rec}}$ with 
KL divergence for both style embedding and content 
embedding~\cite{kingma2013auto}.

The main designs of style- and content-oriented losses are as 
follows~\cite{john2018disentangled}.
\begin{enumerate}
	\item The style embedding should contain enough information to be discriminative. Therefore, a multitask discriminator is added to align the predicted distribution and the ground-truth distribution of labels.
	\begin{align*}
	J_{\mathrm{mul}}(\theta_E; \theta_{\mathrm{mul(s)}}) = -\sum_{l\in \mathrm{labels}}t_s(l)\log y_s(l),
	\end{align*}
	where $t_s(l)$ is the distribution of ground-truth style labels, and $y_s(l)$ is the predicted output by the style discriminator.
	\item The content embedding should not contain too much style information. Therefore, an adversarial discriminator is added, with loss of the discriminator and adversarial loss for the autoencoder given by
	\begin{align*}
	J_{\mathrm{dis}(s)}(\theta_{\mathrm{dis}(s)}) & = - \sum_{l\in \mathrm{labels}} t_s(l)\log y_s(l), \\
	J_{\mathrm{adv(s)}}(\theta_E) & = -\sum_{l\in \mathrm{labels}}y_s(l)\log y_s(l),
	\end{align*}
	where $\theta_{\mathrm{dis}(s)}$ contains the weights for a fully connected layer, and $t_c(l)$ is the predicted distribution of style labels when taking content embedding as an input. 
	\item The content embedding needs to be able to predict the information given by bag-of-words (BoW), which is defined as
	\begin{align*}
	t_c(w) := \frac{\sum_{i=1}^N \mathbb{I}\{w_i = w \}}{N},
	\end{align*}
	for each word $w$ in the vocabulary $V$ with sentence length $N$~\cite{wallach2006topic}. Therefore, a multitask discriminator is added to align the predicted BoW distribution with ground-truth.
	\begin{align*}
	J_{\mathrm{mul}(c)}(\theta_E; \theta_{\mathrm{mul}(c)}) & = -\sum_{w\in V}t_c(w)\log y_c(w),
	\end{align*}
	where $t_c(w)$ is the distribution of true BoW representations, and $y_c(w)$ is the predicted output by the content discriminator.
	\item The style embedding should not contain content information. Similar as before, an adversarial discriminator is trained to predict the BoW features from style embedding, with loss for discriminator and adversarial loss given by
	\begin{align*}
	J_{\mathrm{dis}(c)}(\theta_{\mathrm{dis}(c)}) = -\sum_{w\in V}t_c(w)\log y_c(w), \\
	J_{\mathrm{adv}(c)}(\theta_E) = -\sum_{w\in V}y_c(l)\log y_c(l).
	\end{align*}
\end{enumerate}

In the training phase, the adversarial discriminators are trained 
together with other parts of the model, and the final loss of 
the autoencoder is given by the weighted sum of the loss from traditional VAE, 
the multitask losses for style and content, 
and the adversarial losses given by the style and content discriminators. 
Then in the inference phase, the style embedding is extracted from the 
latent space of a target domain, and the original style embedding is 
substituted by this target embedding in decoding.

\subsection{Model-Agnostic Meta-learning (MAML)}

Meta-learning is designed to help a model quickly adapt to a new tasks, given that it has been trained on several other similar tasks. Compared with other model-based meta-learning methods, model-agnostic meta-learning algorithm (MAML) utilizes only gradient information~\cite{finn2017model}. Therefore, it can be easily applied to models based on gradient descent training.

Given a distribution of similar tasks $p(\mathcal{T})$, a task-specific loss function $\mathcal{L}_{\mathcal{T}_i}$ and shared parameters $\theta$, we aim to jointly learn a model so that in fine-tuning with the new task, the parameters are well-initialized so that the model quickly converges with fewer epochs and a smaller dataset.

\begin{figure}[th]
	\centering
	\includegraphics[width=7cm]{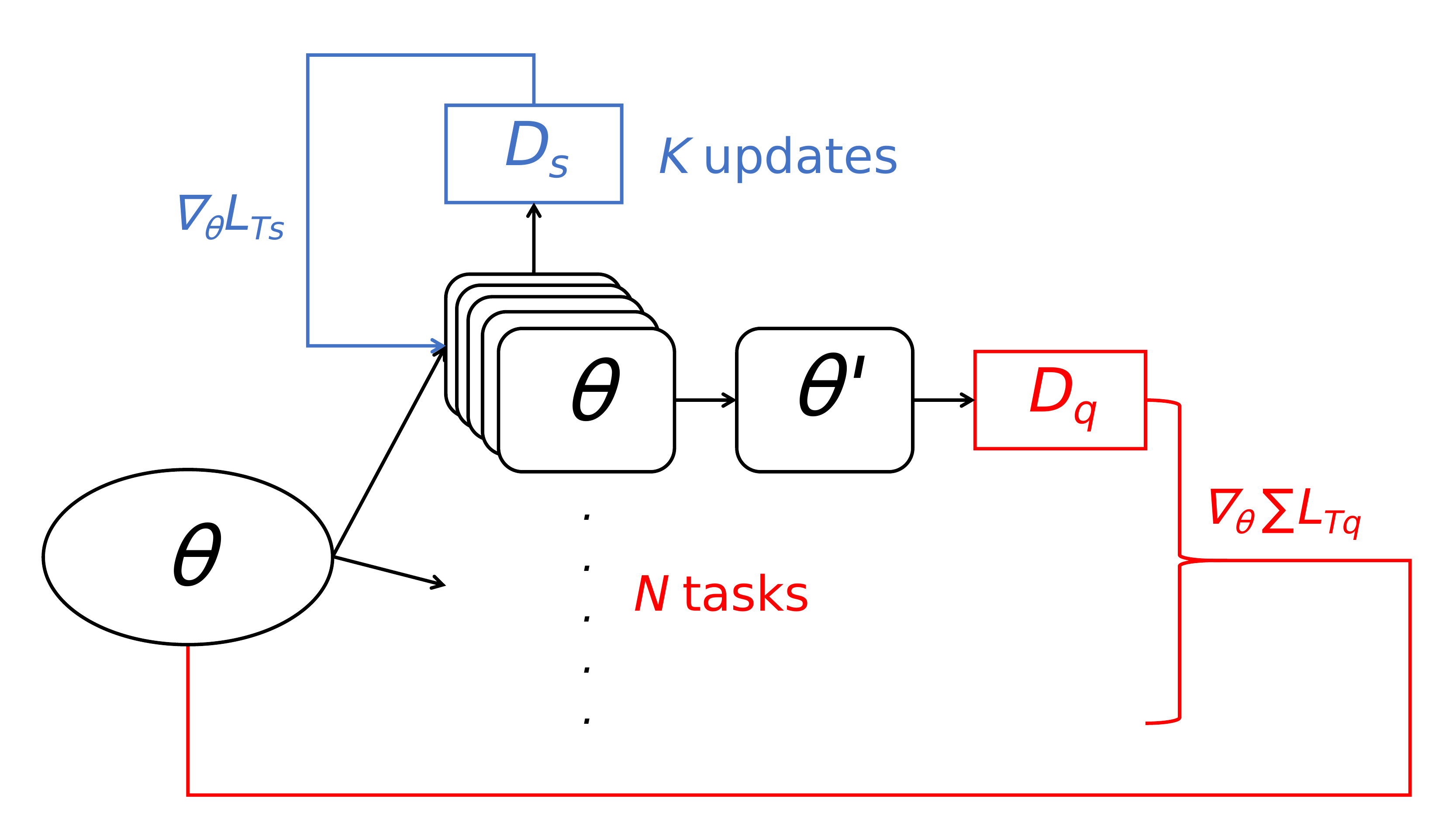}
	\caption{MAML architecture. For every update of meta-learner's parameters $\theta$, we first update each sub-task on the support dataset $D_s$ for $K$ steps and obtain the new parameter $\theta'$. Then we use the loss evaluated using this new parameter on the query set $D_q$, and sum up all losses from $N$ tasks to update meta-learner's parameters.}
	\label{fig:maml}
\end{figure}

Figure \ref{fig:maml} shows the architecture of MAML. We define the shared model with parameters $\theta$ as a meta-learner. The data for each task is divided in to a support set $D_s$ and a query set $D_q$. Every update of the meta-learner's parameters consist of $K$-step updates for each of the $N$ tasks. The support set for each task is used to update the $N$ sub-tasks, and the query set is used to evaluate a query loss that is later used for meta-learner's updates.

In each sub-task training, the sub-learner is initialized with the parameters of the meta-learner. Then this parameter is updated $K$ times using the support data for this specific task. After updating, the new parameter is $\theta'_i$ for the $i$-th task, and a loss $\mathcal{L}_{\mathcal{T}_i}(f_{\theta'})$ is evaluated using the query dataset for this sub-task. This sub-training process is performed for each sub-task, and losses from all sub-tasks are aggregated to obtain a loss for meta-training.

\begin{algorithm}\small
	\caption{ST$^2$}
	\label{alg:maml}
	\KwIn{a set of style pairs, $\{(s_{t,1}, s_{t,2}), \ldots \}$, where $t = 1, \ldots, N$, parameters $\alpha, \beta$}
	\KwOut{transfer function $f_{\theta}: (x, s) \mapsto y$, where $s$ is the source style, $x$ is the original sentence, $y$ is the transferred sentence in target style}
	\While{not done}{
		\ForEach{style pair $(s_{t,1}, s_{t,2})$}{
			Initialize sub learner with $\theta_t = \theta$\;
			\For{step in 1, \ldots, K}{
				Sample batch data from support set of $t$\;
				Update transfer function $f_{\theta}$ using\\$\qquad\theta_t = \theta_t - \alpha \nabla_{\theta_t}\mathcal{L}_{t}(f_{\theta_t})$\;
			}
			Sample batch data from query set of $t$\;
			Evaluate $\mathcal{L}_t(f_{\theta_t})$\;
		}
		Update meta-learner with $\theta = \theta - \beta\nabla_{\theta}\sum_{t=1}^T \mathcal{L}_t(f_{\theta_t})$\;
	}
\end{algorithm}

In our application, the sub-tasks contain different pairs of styles to 
be transferred. The meta-learner contains the transfer function 
$f_{\theta}: (x, s)\mapsto x'$, which takes a sentence $x$ with its 
style label $s$, and outputs a sentence $x'$ in the target style with 
similar content. This transfer function is shared by all pairs of styles 
in the meta-training phase. In addition, both our base models include adversarial functions for style disentanglement, the updates for the adversarial parameters are also included in the updates of meta-learner. Since the data size for each task with a 
single pair of styles is assumed to be small, 
the goal of MAML is to use information from other style pairs for 
a better initialization in the fine-tuning phase of a specific sub-task. 
The multi-task style transfer via meta-learning (ST$^2$) algorithm is 
described in Algorithm \ref{alg:maml}. 

\section{Experiments}
\label{sec:eval}

In order to incorporate a more diverse range of styles, we gather two datasets for our experiments. The first is collected from literature translations with different writing styles, and the second is a grouped standard dataset used for existing style transfer works, which also contains different types of styles.

We test our ST$^2$ model and state-of-the-art models on these two datasets, and verify our model's effectiveness on few-shot style transfer scheme. By comparing our models with the pretrained base models, we verify that meta-learning framework improves the performance both in terms of language fluency and style transfer accuracy.

\begin{table*}[th]\footnotesize
	\centering
	\begin{tabular}{c|cc}
		\hline
		\textbf{Common Source} & \textbf{Writer A} & \textbf{Writer B} \\
		\hline
		Notre-Dame de Paris & Alban Kraisheimer & Isabel F. Hapgood \\
		The Brothers Karamazov & Andrew R. MacAndrew & Richard Pevear \\
		The Story of Stone & David Hawkes & Yang Xianyi \\
		The Magic Mountain & John E. Woods & H. T. Lowe-Porter \\
		The Iliad & Ian C. Johnston & Robert Fagles \\
		Les Miserables & Isabel F. Hapgood & Julie Rose \\
		Crime and Punishment & Michael R. Katz & Richard Pevear \\
		\hline
	\end{tabular}
	\caption{Literature translations dataset. The first column shows the name of translated works with common source for the two writers in the same row.}\label{tb:translations}
\end{table*}

\subsection{Datasets}

Since we extend the definition of style to the general writing style of a person, we do not need to be limited to the widely used Yelp/Amazon review and GYAFC datasets. To model the real situation where we have different style pairs with not enough data for each style pair, we propose to use the literature translations dataset and a set of popular style transfer datasets with reduced sizes.

\subsubsection*{Literature Translations (LT)}
\label{sec:lt}

Current state-of-the-art works on text style transfer require large datasets for training, and thus they are not able to be applied to personal writing styles. One reason is that personal writing styles are relatively difficult to learn, compared with more discriminative styles such as sentiment and formality. Furthermore, sources of data reflecting personal writing styles are quite limited. 

For the reasons above, we consider literature translations dataset. Firstly, there are multiple versions of translation from the same source. Since it is possible to align these comparable sentences to construct ground-truth references, they are well-suited for our test data. Moreover, in addition to the common-source translated work, a translator has other written works, which can be used for our non-parallel training data.

We collect a set of writers with unknown different writing styles $\{s_1, \ldots, s_n\}$, with each writer has his/her own set of written works $\{c_{s_i, 1}, \ldots, c_{s_i, n_i} \}$. In order to have a test set with ground-truth references, we used translated works from non-English sources\footnote{Obtained from \texttt{http://gen.lib.rus.ec/}.}, so that each writer in our set has at least one translated work that is from the same source as another writer. Namely, for each writing style $s_i$ in the set, there exists another style $s_j$ and $\exists\ k_1, k_2$ such that $src(c_{s_i, k_1}) = src(c_{s_j, k_2})$. In this dataset, each writer has approximately 10k nonparallel sentences for training.

We used the aligned sentences for each style pair using the algorithm provided by \citet{chen2019align} for testing. The sentence pairs are extracted from the common translated work for each writer pair. The test data has approximately 1k sentences for each writer. More information is shown in Table \ref{tb:translations}.

\subsubsection*{Grouped Standard Datasets (GSD)}

In our second set, we group popular datasets for style transfer. For large datasets, we use only a small portion of then in order to model our few-shot style transfer task. The datasets we use are listed in Table \ref{tb:data2}. For the standard/simple versions of Wikipedia, we use the aligned sentences by \citet{hwang2015aligning} for testing. For all datasets listed in the table, we use 10k sentences for training and 1k sentences for testing.

\begin{table}[th]\footnotesize
	\centering
	\begin{tabular}{cc}
		\hline
		\textbf{Dataset} & \textbf{Style} \\
		\hline
		Yelp & (health) positive/negative \\
		Amazon & (musical instrument) positive/negative \\
		GYAFC & (relations )formal/informal \\
		Wikipedia & standard/simple \\
		Bible & standard/easy \\
		Britannica & standard/simple \\
		Shakespeare & original/modern \\
		\hline
	\end{tabular}
	\caption{Grouped dataset.}\label{tb:data2}
\end{table}

\subsection{Metrics}
\subsubsection*{BLEU for Content Preservation}

To evaluate content preservation of transferred sentences, we use a multi-BLEU score between reference sentences and generated sentences~\cite{papineni2002bleu}. When ground-truth sentences are available in the dataset, we calculate the BLEU scores between generated sentences and ground-truth sentences. When they are missing, we calculate self-BLEU scores based on the original sentences\footnote{We use BLEU score provided by \texttt{multi-bleu.perl}}.

\subsubsection*{Perplexity (PPL)}

Following the metrics used by \citet{john2018disentangled}, we use a bigram Kneser-Key bigram language model to evaluate the fluency and naturalness of generated sentences~\cite{kneser1995improved}. The language models are trained in the target domain for each style pair. We use the training data before reduction to train the language model for GSD set.

\subsubsection*{Transfer Accuracy (ACC)}

To evaluate the effectiveness of style transfer, we pretrain a TextCNN classifier proposed by \citet{kim2014convolutional}. The transfer accuracy is the score output by the CNN classifier. Our classifier achieves accuracy of 80\% on GSD and 77\% even on LT dataset, which serves as a reasonable evaluator for transfer effectiveness.

\subsubsection*{Human Evaluation of Fluency and Content}

\begin{table*}[th]\footnotesize
	\centering
	\begin{tabular}{c|ccccc|ccccc}
		\hline
		\multirow{2}{*}{\textbf{Model}} & \multicolumn{5}{c|}{\textbf{LT}} & \multicolumn{5}{c}{\textbf{GSD}} \\
		\cline{2-11}
		& \textbf{B-ref}$^{\uparrow}$ & \textbf{B-ori} & \textbf{PPL}$^\downarrow$ & \textbf{ACC}$^\uparrow$ & \textbf{Human}$^\uparrow$ & \textbf{B-ref}$^\uparrow$ & \textbf{B-ori} & \textbf{PPL}$^\downarrow$ & \textbf{ACC}$^\uparrow$ & \textbf{Human}$^\uparrow$ \\
		\hline
		Template & \textbf{41.6} & 81.48 & \textbf{5.4} & 0.31 & \textbf{4.3} / \textbf{4.2} & \textbf{81.7} & 88.8 & \textbf{5.3} & 0.42 & 4.2 / \textbf{4.2} \\
		\underline{CrossAlign} & 2.2 & 2.1 & 1895.6 & 0.45 & 1.2 / 1.1 & 2.7 & 2.2 & 1049.7 & 0.36 & 1.0 / 1.0 \\
		DeleteRetrieve & \textbf{35.9} & 41.6 & 63.3 & 0.33 & 1.0 / 1.0 & 20.5 & 21.4 & 28.8 & 0.41 & 2.1 / 1.3 \\
		DualRL & 4.1 & 3.9 & 1400.7 & 0.49 & 1.2 / 1.2 & 25.4 & 27.5 & 171.0 & 0.41 & 2.9 / 1.5 \\
		\underline{VAE} & 13.5 & 16.3 & 8.5 & 0.49 & 3.5 / 1.7 & 12.4 & 26.4 & 21.5 & 0.45 & \textbf{4.3} / 2.1 \\
		\hline
		ST$^2$-CA (ours) & 6.3 & 6.8 & 54.8 & \textbf{0.65} & 3.1 / \textbf{2.3} & \textbf{66.7} & 73.2 & 21.4 & 0.42 & 3.6 / \textbf{3.8} \\
		ST$^2$-VAE (ours) & 20.5 & 15.1 & \textbf{8.2} & 0.62 & \textbf{3.8} / 1.9 & 14.7 & 13.9 & \textbf{10.9} & \textbf{0.71} & \textbf{4.3} / 2.7 \\
		\hline
	\end{tabular}
	\caption{Results for multi-task style transfer. The larger$^\uparrow$/lower$^\downarrow$, the better. B-ref and B-ori means BLEU score and self-BLEU score, respectively. The human evaluation scores include language fluency/content preservation, respectively. Our base models are underlined.}\label{tb:exp1}
\end{table*}

We conduct an additional human evaluation, following \citet{luo2019dual}. Two native English speakers are required to score the generated sentences from 1 to 5 in terms of fluency, naturalness, and content preservation, respectively. Before annotation, the two evaluators are given the best and worst sentences generated so as to know the upper and lower bound, and thus score more linearly. The final score for each model is calculated as the average score given by the annotators. The kappa inter-judge agreement is 0.769, indicating significant agreement. 

\subsection{Multi-task Style Transfer}
\label{sec:st}

We compare the results with the state-of-the-art models for the style transfer task. All the baseline models are trained on the single style pair. The ST$^2$ model is trained on all the tasks for both LT and GSD sets, and then fine-tuned using a specific style pair in the sets. The trained meta-learner is fine-tuned on each of the sub-tasks, and the scores are calculated as the average among all sub-tasks for both ST$^2$ models and baselines. The results are shown in Table \ref{tb:exp1}. 

We note that the BLEU and PPL scores for the template based model appear
to be superior to those of other models. This is because it directly 
modifies the original sentence by changing a couple of words. So the modification
is actually minimum. However, its transfer accuracy suffers, which is
well expected. Thus its should only serve as a reference in our task.

For qualitative analysis, we randomly select sample sentences output 
by the baseline models, pretrained base models and our ST$^2$ models 
on the Translations dataset and Yelp positive/negative review dataset, 
which are shown in Table \ref{tb:qual}.

\begin{table*}[th]\footnotesize
	\centering
	\begin{tabular}{c|c}
		\hline
		\textbf{Original Sentence} (Notre-Dame de Paris) & \emph{in their handsome tunics of purple camlet , with big white crosses on the front .} \\
		\hline
		Template & \emph{in their handsome tunics of purple camlet, with big white crosses on front.} \\
		CrossAlign & \emph{heel skilful skilful skilful skilful} \\
		DeleteRetrieve & \emph{the man , and the man , the man , the} \\
		DualRL & \emph{lyres lyres lyres} \\
		VAE & \emph{the gypsy girl had stirred up from the conflict} \\
		\hline
		ST$^2$-CrossAlign (ours) & \emph{so the spectacle who prayed and half white streets ,} \\
		ST$^2$-VAE (ours) & \emph{all four were dressed in robes of white and were white locks from} \\
		\hline
		\hline
		\textbf{Original Sentence} (Yelp positive) & \emph{the staff is welcoming and professional .} \\
		\hline
		Template & \emph{the staff is welcoming and professional .} \\
		CrossAlign & \emph{glad glad glad} \\
		CrossAlign (pretrained) & \emph{the staff is welcoming and professional .} \\
		DeleteRetrieve & \emph{the staff is a time .} \\
		DualRL & \emph{less expensive have working .} \\
		VAE & \emph{the staff is rude and rude} \\
		VAE (pretrained) & \emph{the staff is extremely welcoming and professional .} \\
		\hline
		ST$^2$-CrossAlign (ours) & \emph{the staff is friendly and unprofessional} \\
		ST$^2$-VAE (ours) & \emph{the staff are rude and unprofessional .} \\
		\hline
		\hline
		\textbf{Original Sentence} (Yelp negative) & \emph{these people do not care about patients at all !} \\
		\hline
		Template & \emph{these people wonderful about patients at all !} \\
		CrossAlign & \emph{glad glad glad} \\
		CrossAlign (pretrained) & \emph{these people do not care about patients at all !} \\
		DeleteRetrieve & \emph{i was n't be a a appointment and i have .} \\
		DualRL & \emph{and just like that it was over and i was .} \\
		VAE & \emph{these people do not care about patients or doctors} \\
		VAE (pretrained) & \emph{these guys do n't care about the patients at time} \\
		\hline
		ST$^2$-CrossAlign (ours) & \emph{these people do not satisfied at all !} \\
		ST$^2$-VAE (ours) & \emph{i was so happy and i did n't consent} \\
		\hline
	\end{tabular}
	\caption{Randomly selected sample outputs for the Alban Kraisheimer/Isabel F. Hapgood pair in LT dataset and Yelp positive/negative review dataset.}\label{tb:qual}
\end{table*}

From the results, we notice that state-of-the-art models fail to achieve satisfying performances in our few-shot style transfer task, and many baseline models fail to generate syntactically or logically consistent sentences. In comparison, our methods are able to generate relatively fluent sentences both in terms of automatic evaluation and human evaluation, meanwhile achieving a higher transfer accuracy.

We might be tempted to conclude that this is simply because the 
ST$^2$ models learn better language models because they are trained on larger data, 
i.e., data from all styles rather than only a single pair of styles. 
Therefore, further experiments are required.

\subsection{Pretrained Base Models}
\label{sec:pretrain}

Based on the previous reasoning, we extract and pretrain the language model part in our base models (\emph{CrossAlign} and \emph{VAE}) on the union of data from all sub-tasks. Starting with a well-trained language model, we then fine-tune the models for the style transfer task. By comparing these models with our ST$^2$ model, we verify that meta-learning framework can improve the style transfer accuracy in addition to language fluency. We perform this experiment only on the GSD dataset, since they are enough for analysis purposes. 

In addition, to examine the effect of pretraining combined with meta-learning, we also add a pretraining phase to our ST$^2$ model. The quantitative and qualitative results are included in Table \ref{tb:exp2} and Table \ref{tb:qual} (on Yelp dataset for the pretrained base models).

By adding a pretraining phase, the models get a chance to see all the data and learn to generate fluent sentences via reconstruction. Therefore, it is not surprising that the content preservation measure (BLEU) and sentence naturalness measure (PPL) give significantly better results than before but at a cost of style transfer accuracy. 

In effect, the models tend to reconstruct the original sentence and do not transfer the style. In comparison, our ST$^2$ model learn to generate reasonable sentences and transfer styles jointly in the training phase. Therefore, it is still superior in terms of style transfer accuracy. This verifies that the success of ST$^2$ is not merely resulted from a larger training dataset. The way that the model updates its knowledge is parallel, rather than sequential, which contributes to better language models and more effective style transfer.

Furthermore, we notice that the pretraining phase in our ST$^2$ model is not crucial, suggesting that it is the meta-learning framework that significant contributes to the model's improvements in generating fluent sentences and effectively transferring styles.

\begin{table}[th]\footnotesize
	\centering
	\begin{tabular}{c|cccc}
		\hline
		\textbf{Model} & \textbf{BLEU}$^\uparrow$ & \textbf{PPL}$^\downarrow$ & \textbf{ACC}$^\uparrow$ & \textbf{Human}$^\uparrow$ \\
		\hline
		CA (pre.) & \textbf{70.4} & 12.2 & 0.32 & 3.9 \\
		VAE (pre.) & 17.2 & 22.4 & 0.48 & 4.0 \\
		\hline
		ST$^2$-CA (pre.) & 62.7 & 23.2 & 0.37 & 3.7 \\
		ST$^2$-VAE (pre.) & 13.6 & 10.9 & 0.66  & 4.2 \\
		\hline
		ST$^2$-CA (ours) & 66.7 & 21.4 & 0.42 & 3.6 \\
		ST$^2$-VAE (ours) & 14.7 & \textbf{10.9} & \textbf{0.71} & \textbf{4.3} \\
		\hline
	\end{tabular}
	\caption{Results on GSD for pretrained (pre.) base models (CrossAlign abbreviated as CA) and ST$^2$.}\label{tb:exp2}
\end{table}

\subsection{Disentanglement of Style}
\label{sec:disentangle}

\begin{figure}[th]
	\underline{\small Pretrained CrossAlign}\\
	\begin{minipage}{0.45\linewidth}
		\centering
		\includegraphics[width=3.5cm]{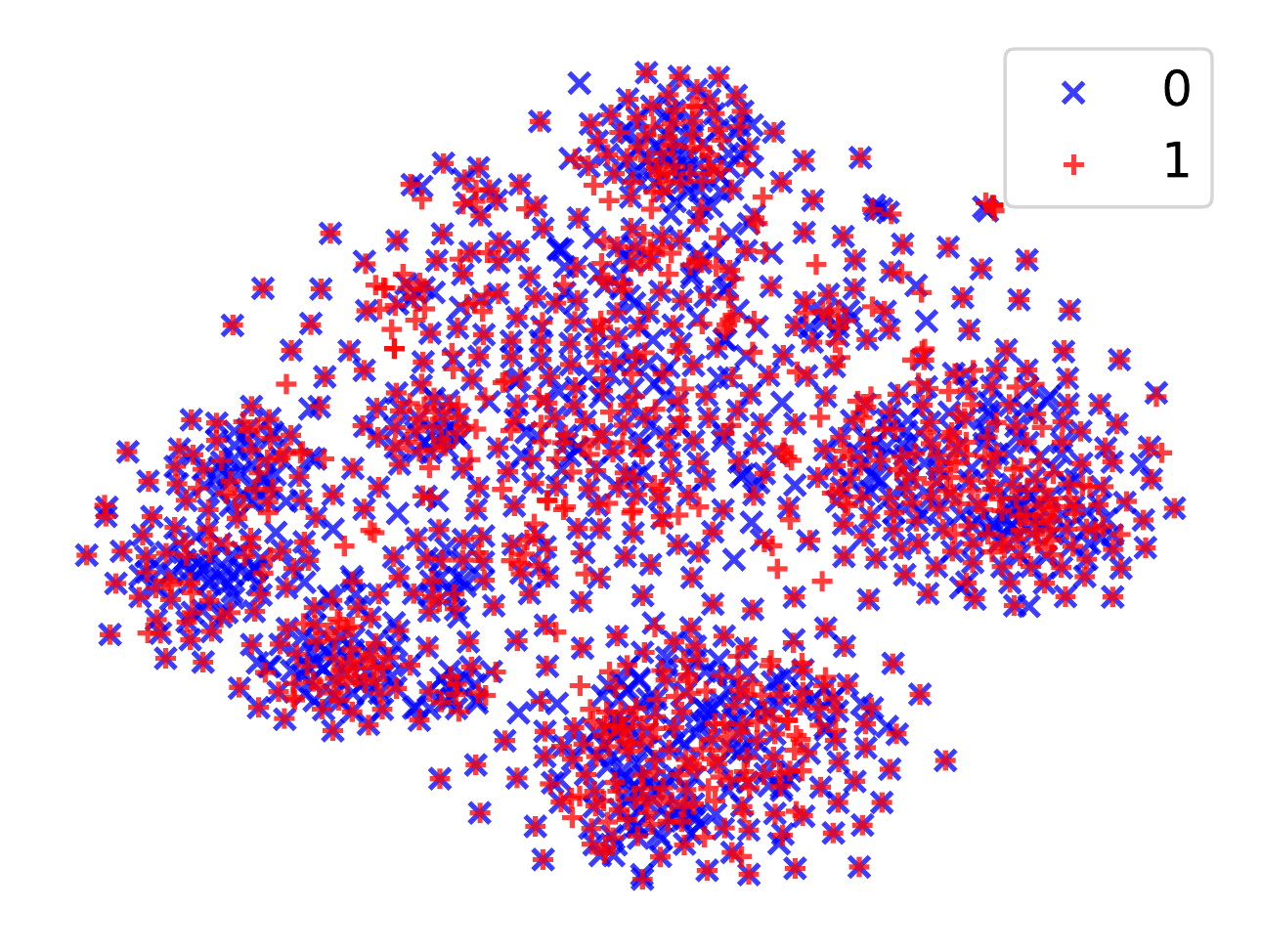}
	\end{minipage}
	\begin{minipage}{0.45\linewidth}
		\centering
		\includegraphics[width=3.5cm]{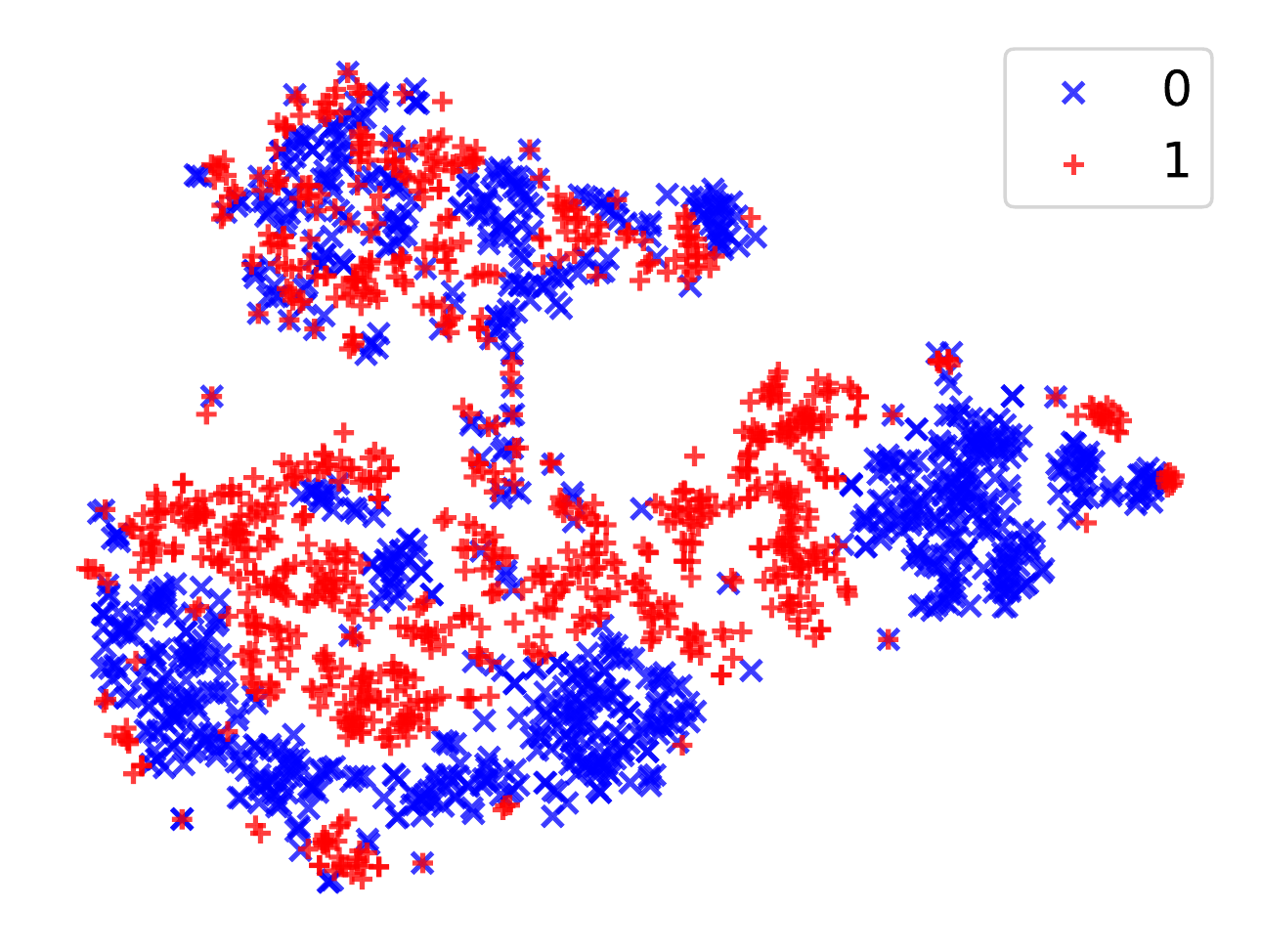}
	\end{minipage}
	\underline{\small ST$^2$-CrossAlign}\\
	\begin{minipage}{0.45\linewidth}
		\centering
		\includegraphics[width=3.5cm]{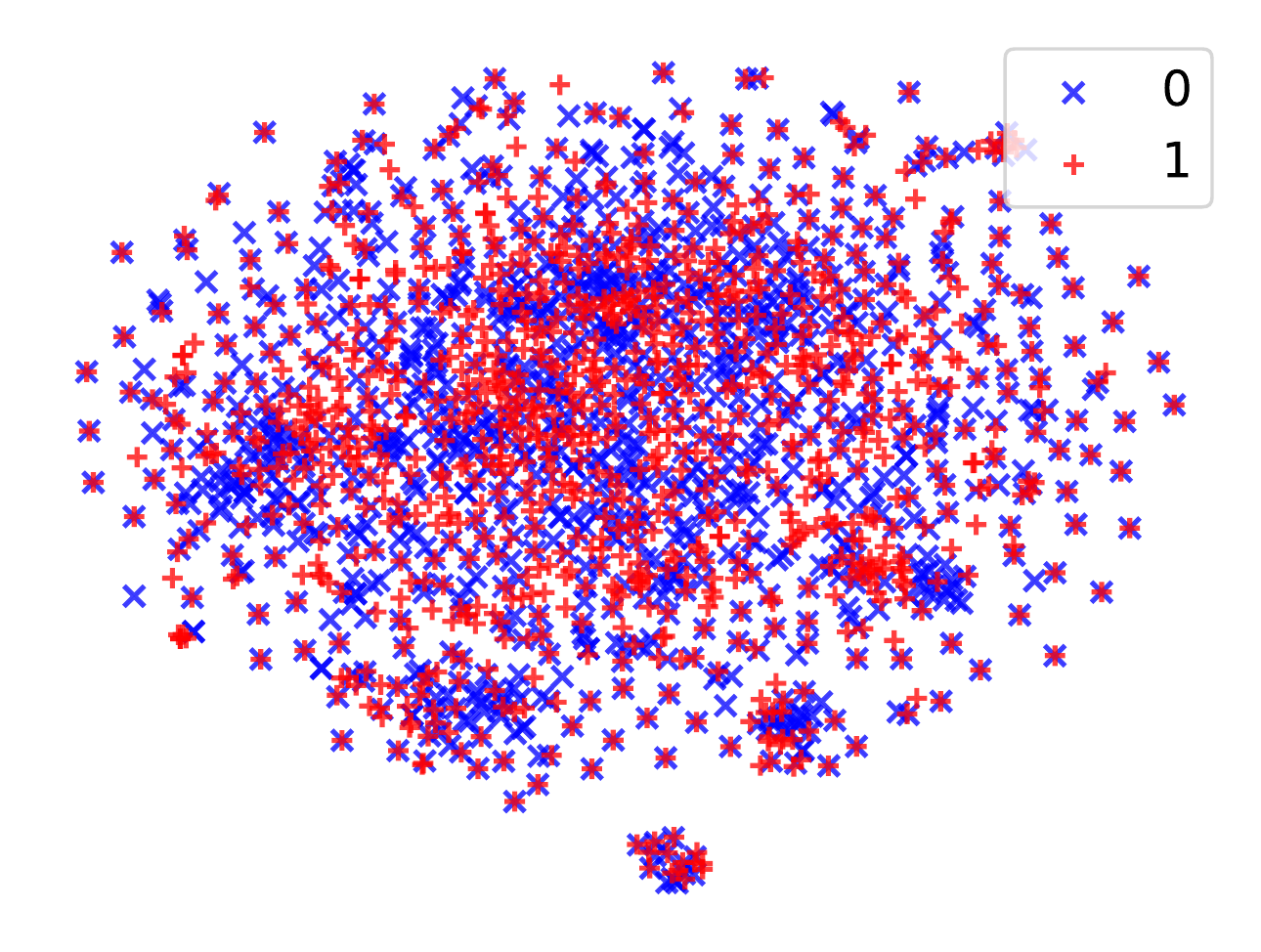}
	\end{minipage}
	\begin{minipage}{0.45\linewidth}
		\centering
		\includegraphics[width=3.5cm]{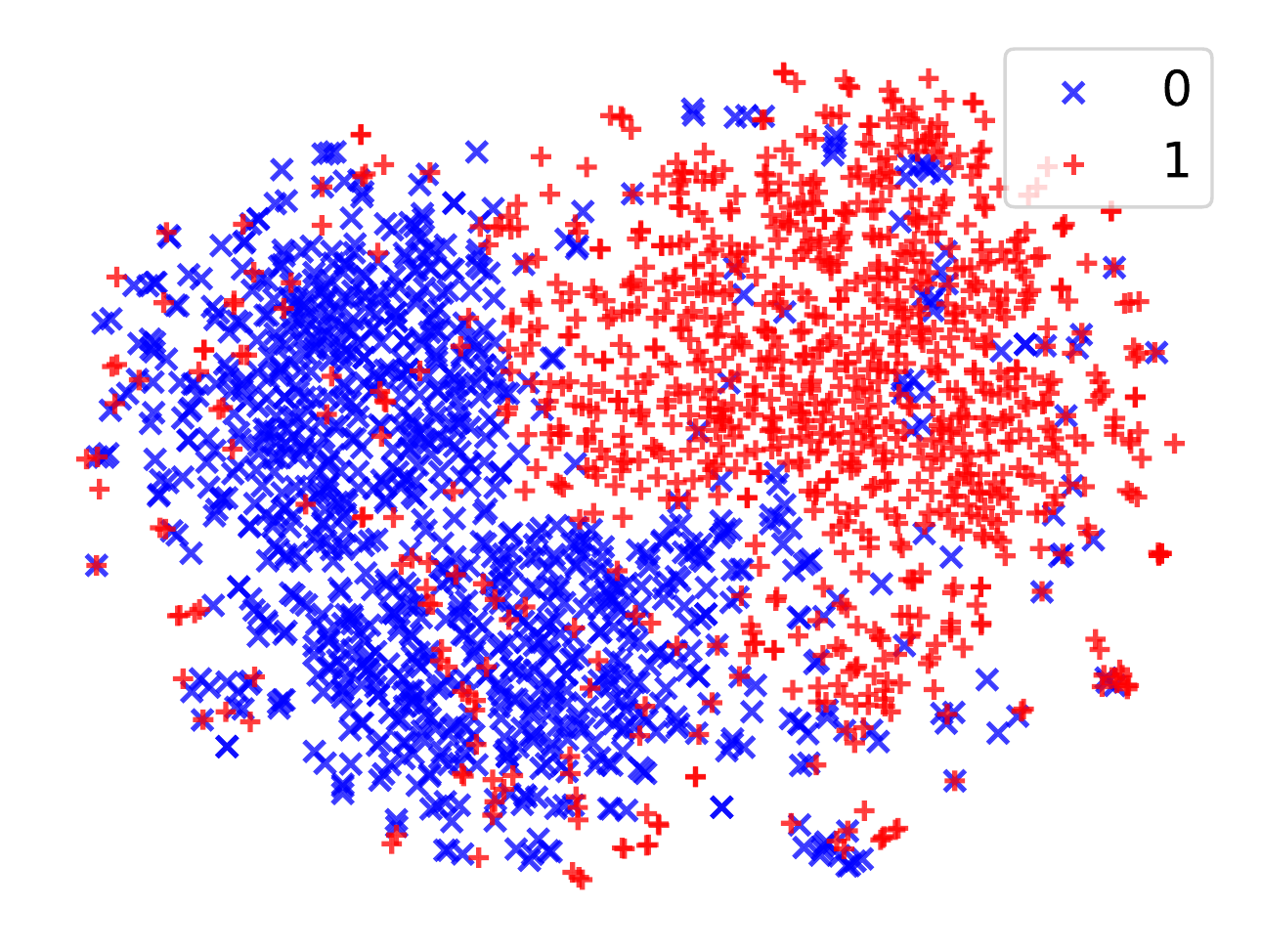}
	\end{minipage}
	\underline{\small Pretrained VAE}\\
	\begin{minipage}{0.45\linewidth}
		\centering
		\includegraphics[width=3.5cm]{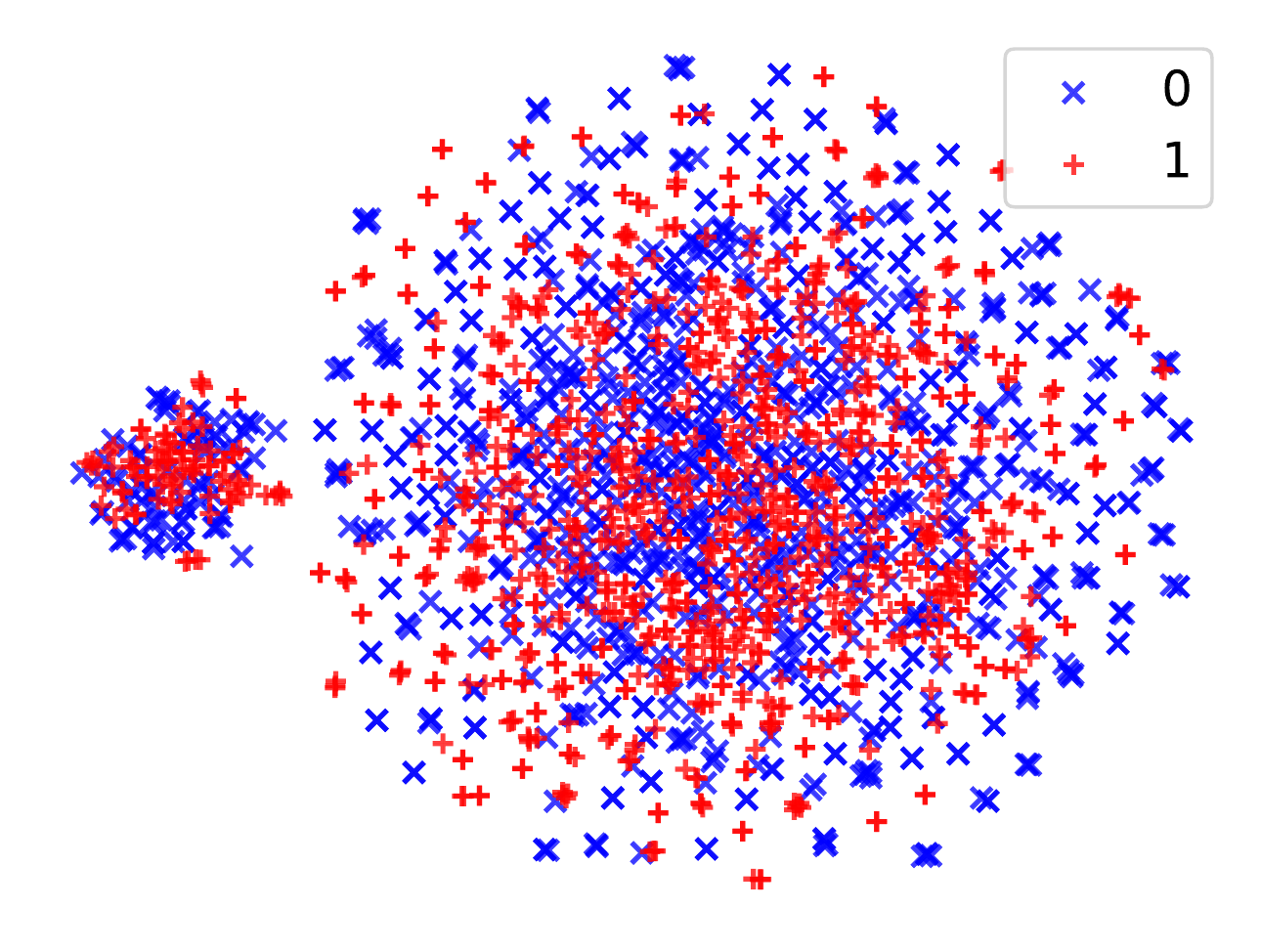}
	\end{minipage}
	\begin{minipage}{0.45\linewidth}
		\centering
		\includegraphics[width=3.5cm]{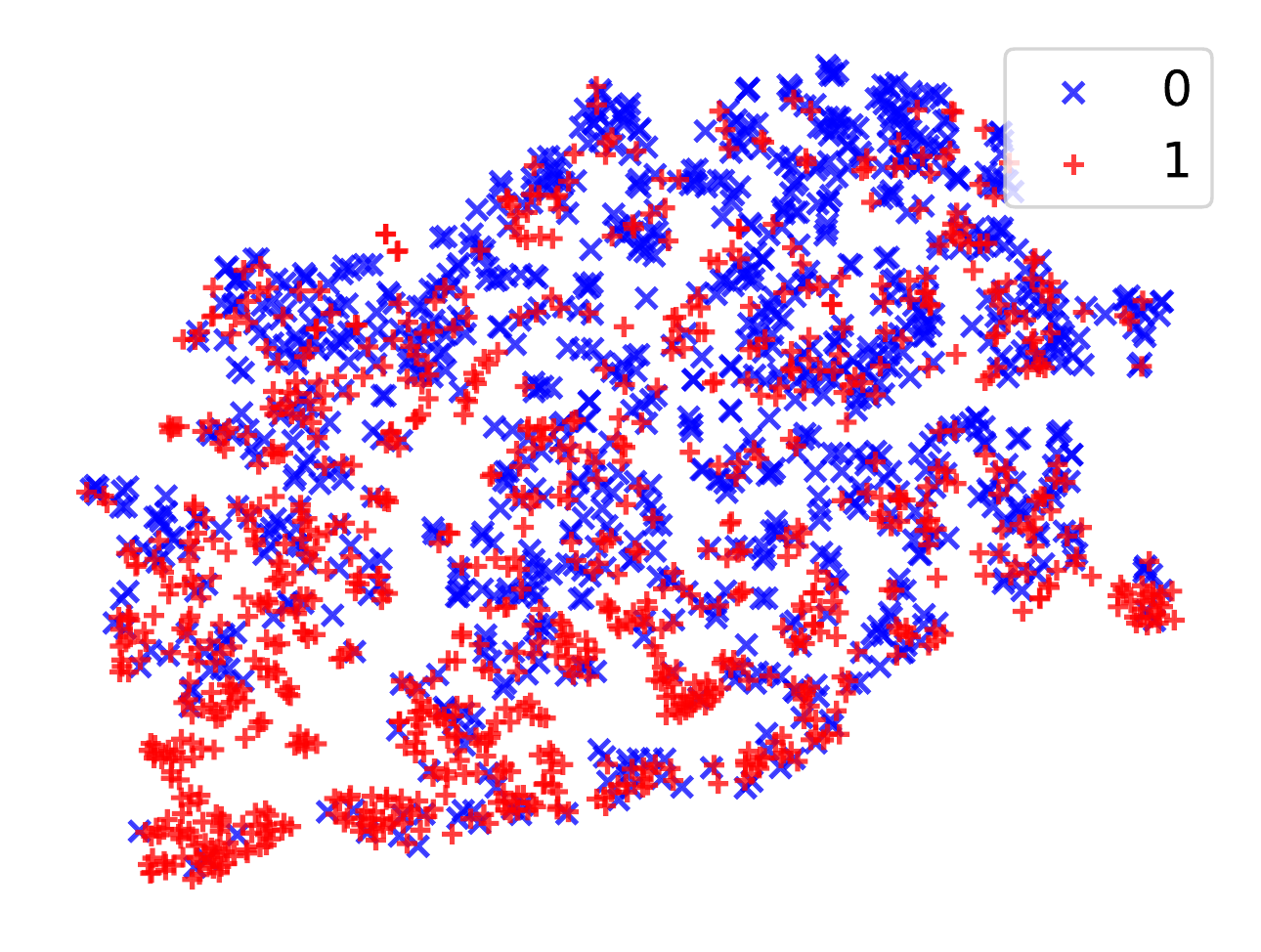}
	\end{minipage}
	\underline{\small ST$^2$-VAE}\\
	\begin{minipage}{0.45\linewidth}
		\centering
		\includegraphics[width=3.5cm]{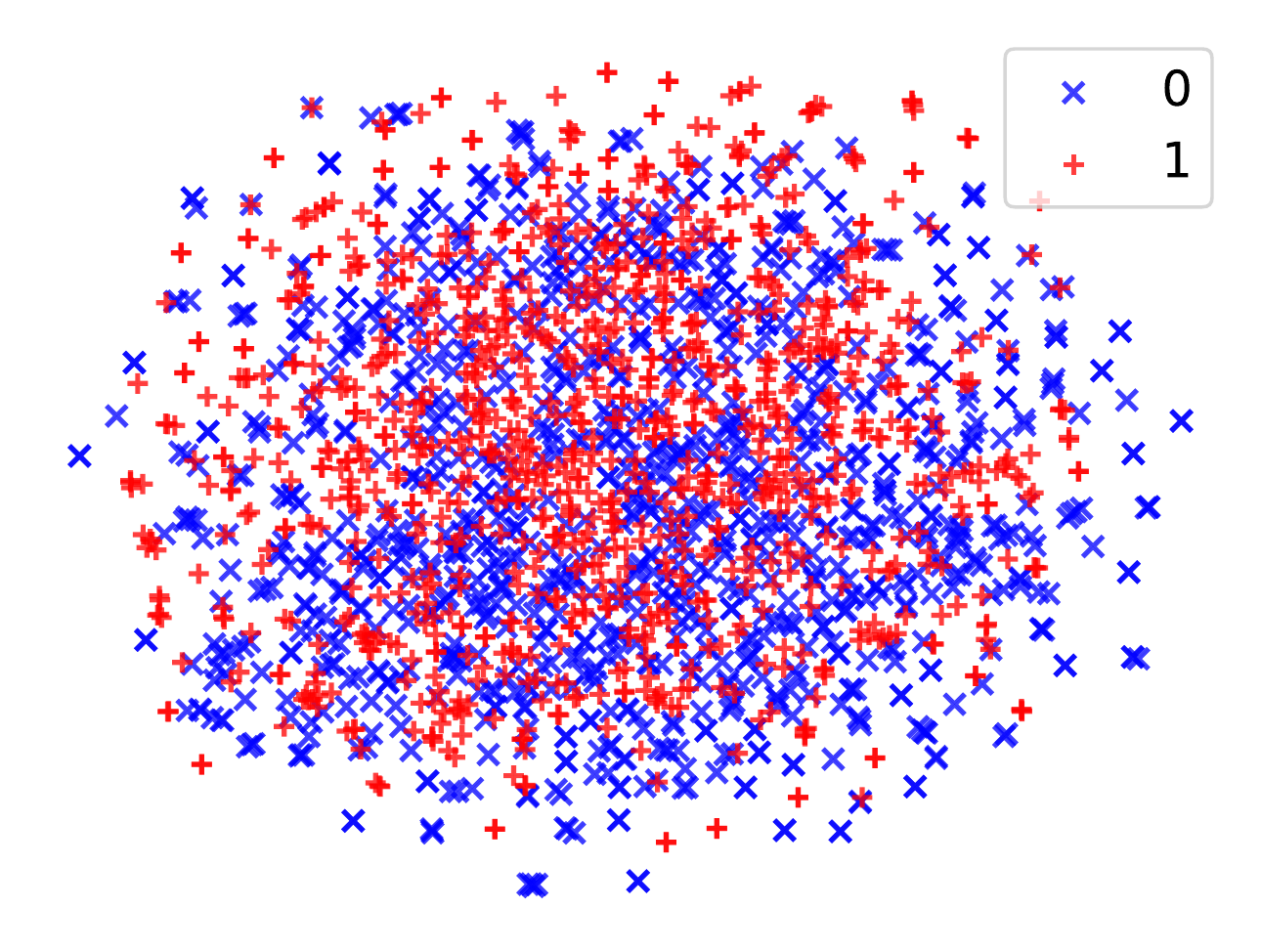}
	\end{minipage}
	\begin{minipage}{0.45\linewidth}
		\centering
		\includegraphics[width=3.5cm]{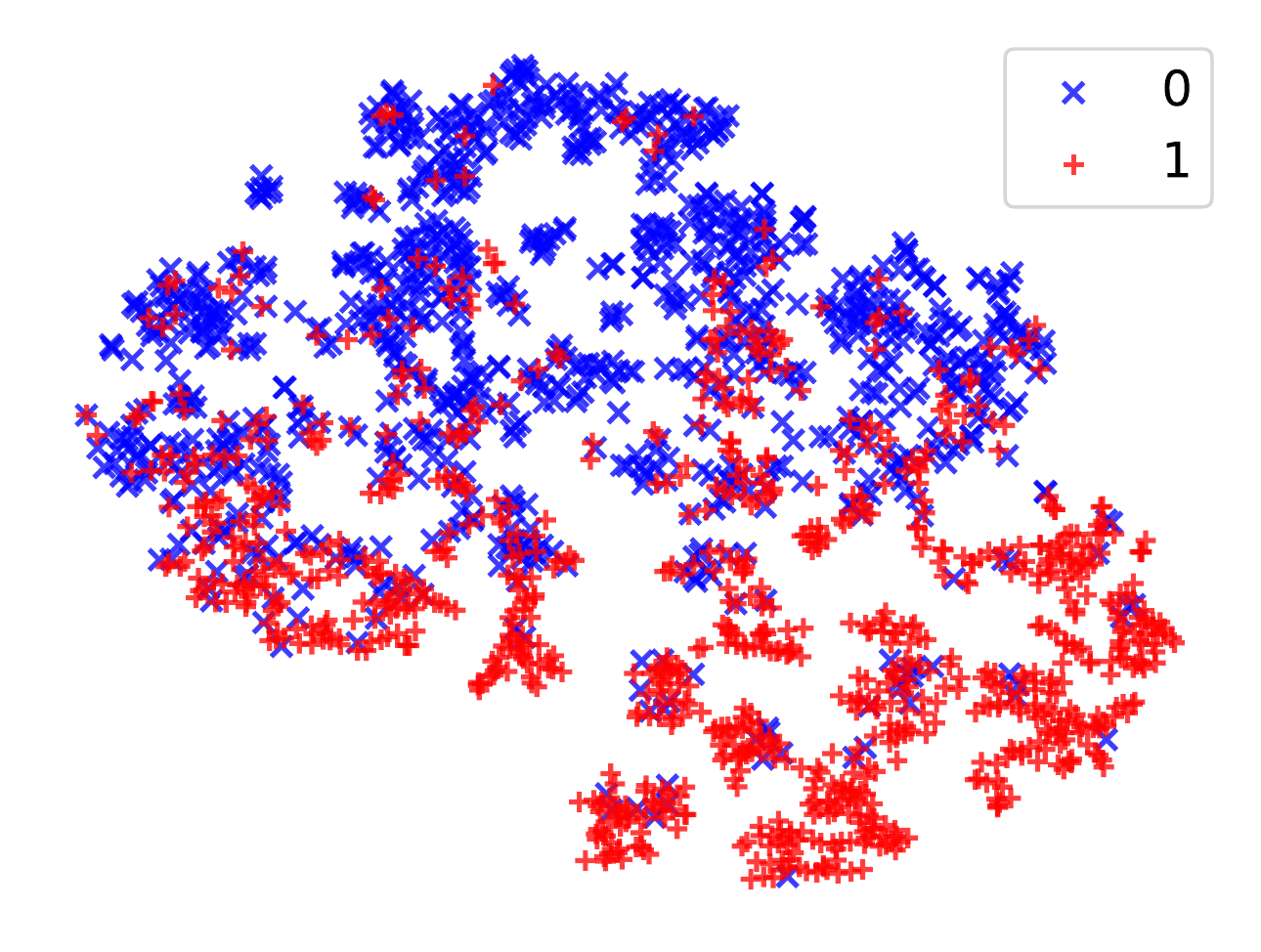}
	\end{minipage}
	\caption{t-SNE plots for content(left) and style(right) embedding}\label{fig:tsne}
\end{figure}

Following the experiments adapted by \citet{john2018disentangled}, we use t-SNE plots (shown in Figure \ref{fig:tsne}) to analyze the effectiveness of disentanglement of style embedding and content embedding in the latent space~\cite{maaten2008visualizing}. In particular, we compare the pretrained base models (\emph{CrossAlign} and \emph{VAE}) and our ST$^2$ models. 

These two models, together with our ST$^2$ models attempt to disentangle style and content in latent space, and thus is well suited for this experiment, while it is unreasonable to treat hidden state vectors in other baseline models as content/style embedding. Therefore, they are excluded from this experiment.

As we can see from the figures, the content space learned by all models are relatively clustered, while the style spaces are more separated in our ST$^2$ models than the pretrained base models. This verifies that the improvements of meta-learning framework is not limited to a better language model, but also in terms of the disentanglement of styles.

\section{Related Work}
\label{sec:related}
\citet{fu2018style} devised a multi-encoder and multi-embedding scheme to learn a style vector via adversarial training. Adapting a similar idea, \citet{zhang2018shaped} built a shared private encoder-decoder model to control the latent content and style space. Also based on a seq2seq model, \citet{shen2017style} proposed a cross-align algorithm to align the hidden states with a latent style vector from target domain using teacher-forcing. More recently, \citet{john2018disentangled} used well-defined style-oriented and content-oriented losses based on a variational autoencoder to separate style and content in latent space.

\citet{li2018delete} directly removed style attribute words based on TF-IDF weights and trained a generative model that takes the remaining content words to construct the transferred sentence. Inspired by the recent achievements of masked language models, \citet{wu2019mask} used an attribute marker identifier to mask out the style words in source domain, and trained a ``infill'' model to generate sentences in target domain.

Based on reinforcement learning, \citet{xu2018unpaired} proposed a cycled-RL scheme with two modules, one for removing emotional words (neutralization), and the other for adding sentiment words (emotionalization). \citet{wu2019hierarchical} devised a hierarchical reinforced sequence operation method using a \emph{point-then-operate} framework, with a high-level agent proposing the position in a sentence to operate on, and a low-level agent altering the proposed positions. \citet{luo2019dual} proposed a dual reinforcement learning model to jointly train the transfer functions using explicit evaluations for style and content as a guidance. Although their methods work well in large datasets such as Yelp~\cite{asghar2016yelp} and GYAFC~\cite{rao2018dear}, it fails in our few-shot style transfer task.

\citet{prabhumoye2018style} adapted a back-translation scheme in an attempt to remove stylistic characteristics in some intermediate language domain, such as French.

There are also meta-learning applications on text generation tasks. \citet{qian2019domain} used the model agnostic meta-learning algorithm for domain adaptive dialogue generation. However, their task has paired data for training, which is different from our task. In order to enhance the content-preservation abilities, \citet{li2019domain} proposed to first train an autoencoder on both source and target domain. But in addition to utilizing off-domain data, we are applying meta-learning method to enhance models' performance both in terms of language model and transfer abilities.

\section{Conclusion}
\label{sec:conclude}

In this paper, we extend the concept of style to general writing styles, 
which naturally exist as many as possible but with a limited size of data. 
To tackle this new problem, we propose a multi-task style transfer (ST$^2$) 
framework, which is the first of its kind to apply meta-learning 
to small-data text style transfer. We use the literature translation dataset 
and the augmented standard dataset to evaluate the state-of-the-art models and 
our proposed model. 

Both quantitative and qualitative results show that ST$^2$ outperforms the state-of-the-art baselines. 
Compared with state-of-the-art models, our model does not rely on a large dataset 
for each style pair, but is able to effectively use off-domain information 
to improve both language fluency and style transfer accuracy.

Noticing that baseline models might not be able to learn an 
effective language model from small datasets, which is a possible reason 
for their bad performances, we further eliminate the bias in our experiment 
by pretraining the base models using data from all tasks. 
From the results, we ascertain that the enhancement of meta-learning framework 
is substantial.

\bibliography{style}
\bibliographystyle{acl_natbib}

\end{document}